\documentclass[10pt,twocolumn,letterpaper]{article}

\usepackage{iccv}
\usepackage{times}
\usepackage{epsfig}
\usepackage{graphicx}
\usepackage{amsmath}
\usepackage{amssymb}
\usepackage{subfig}
\usepackage{floatrow}
\newfloatcommand{capbtabbox}{table}[][\FBwidth]

\DeclareMathOperator*{\argmin}{arg\,min}

\newcommand{\comment}[1]{}

\usepackage[pagebackref=true,breaklinks=true,letterpaper=true,colorlinks,bookmarks=false]{hyperref}

\iccvfinalcopy 

\def\iccvPaperID{894} 

\newcommand\blfootnote[1]{%
	\begingroup
	\renewcommand\thefootnote{}\footnote{#1}%
	\addtocounter{footnote}{-1}%
	\endgroup
}

\ificcvfinal\fi
\begin{document}

\title{SSD-6D: Making RGB-Based 3D Detection and 6D Pose Estimation Great Again }

\author{
	Wadim Kehl $^{1,2,*}$ \hspace{0.3cm} Fabian Manhardt $^{2,*}$ \hspace{0.3cm} Federico Tombari $^{2}$ \hspace{0.3cm} Slobodan Ilic $^{2,3}$ \hspace{0.3cm} Nassir Navab $^{2}$ \\
	\hspace{-0.3cm}
	$^{1}$ Toyota Research Institute, Los Altos  \hspace{0.3cm} $^{2}$ Technical University of Munich  \hspace{0.3cm}
	$^{3}$ Siemens R\&D, Munich \\
	{\tt\small wadim.kehl@tri.global     \hspace{0.5cm}   fabian.manhardt@tum.de  \hspace{0.3cm} tombari@in.tum.de }
}
\maketitle

\begin{abstract}
We present a novel method for detecting 3D model instances and estimating their 6D poses from RGB data in a single shot. To this end, we extend the popular SSD paradigm to cover the full 6D pose space and train on synthetic model data only. Our approach competes or surpasses current state-of-the-art methods that leverage RGB-D data on multiple challenging datasets. Furthermore, our method produces these results at around 10Hz, which is many times faster than the related methods. For the sake of reproducibility, we make our trained networks and detection code publicly available.\footnote{ \url{https://wadimkehl.github.io/} }
\end{abstract}

\blfootnote{* The first two authors contributed equally to this work.}

\vspace*{-0.5cm}

\section{Introduction}
While category-level classification and detection from images has recently experienced a tremendous leap forward thanks to deep learning, the same has not yet happened for what concerns 3D model localization and 6D object pose estimation. 
In contrast to large-scale classification challenges such as PASCAL VOC \cite{Everingham2014} or ILSVRC \cite{Russakovsky2015}, the domain of 6D pose estimation requires instance detection of known 3D CAD models with high precision and accurate poses, as demanded by applications in the context of augmented reality and robotic manipulation. 

Most of the best performing 3D detectors follow a view-based paradigm, in which a discrete set of object views is generated and used for subsequent feature computation \cite{Ulrich2012, Hinterstoisser2012}. During testing, the scene is sampled at discrete positions, features computed and then matched against the object database to establish correspondences among training views and scene locations. Features can either be an encoding of image properties (color gradients, depth values, normal orientations) \cite{Hinterstoisser2012a, Hodan2015, Kehl2015} or, more recently, the result of learning \cite{Brachmann2014, Tejani2014, Brachmann2016, Doumanoglou2016, Kehl2016a}. In either case, the accuracy of both detection and pose estimation hinges on three aspects: 
(1) the coverage of the 6D pose space in terms of viewpoint and scale, (2) the discriminative power of the features to tell objects and views apart and (3) the robustness of matching towards clutter, illumination and occlusion.

CNN-based category detectors such as YOLO \cite{Redmon2016} or SSD \cite{Liu2016} have shown terrific results on large-scale 2D datasets. Their idea is to inverse the sampling strategy such that scene sampling is not anymore a set of discrete input points leading to continuous output. Instead, the input space is dense on the whole image and the output space is discretized into many overlapping bounding boxes of varying shapes and sizes. This inversion allows for smooth scale search over many differently-sized feature maps and simultaneous classification of all boxes in a single pass. In order to compensate for the discretization of the output domain, each bounding box regresses a refinement of its corners. 

The goal of this work is to develop a deep network for object detection that can accurately deal with 3D models and 6D pose estimation by assuming an RGB image as unique input at test time. To this end, we bring the concept of SSD over to this domain with the following contributions: (1) a training stage that makes use of synthetic 3D model information only, (2) a decomposition of the model pose space that allows for easy training and handling of symmetries and (3) an extension of SSD that produces 2D detections and infers proper 6D poses.

We argue that in most cases, color information alone can already provide close to perfect detection rates with good poses. Although our method does not need depth data, it is readily available with RGB-D sensors and almost all recent state-of-the-art 3D detectors make use of it for both feature computation and final pose refinement. We will thus treat depth as an optional modality for hypothesis verification and pose refinement and will assess the performance of our method with both 2D and 3D error metrics on multiple challenging datasets for the case of RGB and RGB-D data.

Throughout experimental results on multiple benchmark datasets, we demonstrate that our color-based approach is competitive with respect to state-of-the-art detectors that leverage RGB-D data or can even outperform them, while being many times faster. Indeed, we show that the prevalent trend of overly relying on depth for 3D instance detection is not justified when using color correctly.

\begin{figure*}
	\includegraphics[width=17.5cm]{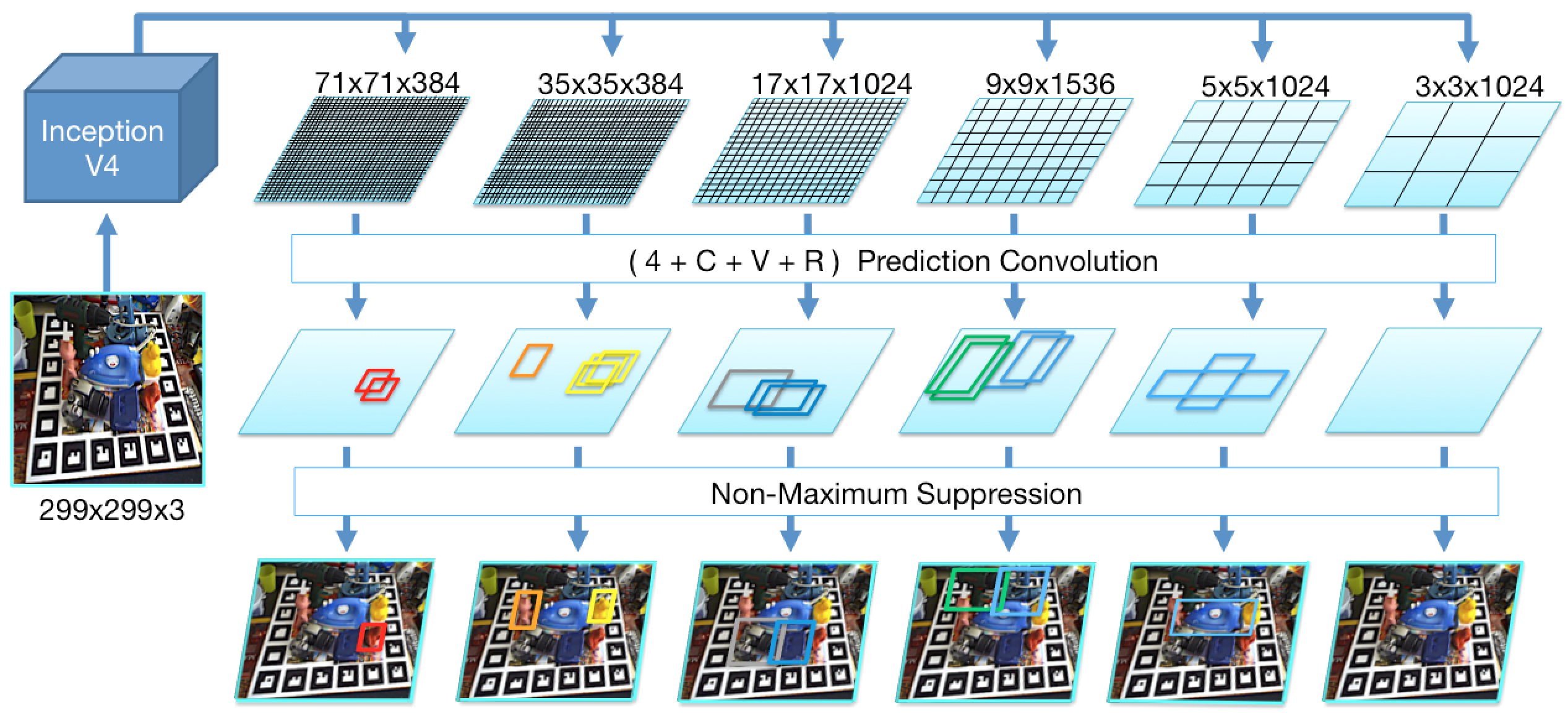}
	\caption{Schematic overview of the SSD-style network prediction. We feed our network with a $299\times299$ RGB image and produce six feature maps at different scales from the input image using branches from InceptionV4.
		Each map is then convolved with trained prediction kernels of shape (4 + C + V + R) to determine object class, 2D bounding box as well as scores for possible viewpoints and in-plane rotations that are parsed to build 6D pose hypotheses. Thereby, C denotes the number of object classes, V the number of viewpoints and R the number of in-plane rotation classes. The other 4 values are utilized to refine the corners of the discrete bounding boxes to tightly fit the detected object.}
	\label{fig:network}
\end{figure*}

\section{Related work}
We will first focus on recent work in the domain of 3D detection and 6D pose estimation before taking a closer look at SSD-style methods for category-level problems.

To cover the upper hemisphere of one object with a small degree of in-plane rotation at multiple distances, the authors in \cite{Hinterstoisser2012} need 3115 template views over contour gradients and interior normals. Hashing of such views has been used to achieve sub-linear matching complexity \cite{Kehl2015, Hodan2015}, but this usually trades speed for accuracy. Related scale-invariant approaches \cite{Hodan2015, Brachmann2014, Tejani2014, Doumanoglou2016, Kehl2016a} employ depth information as an integral part for either feature learning or extraction, thus avoiding scale-space search and cutting down the number of views by around an order of magnitude. Since they require depth to work, they can fail when depth is missing or erroneous. While scale can be inferred with RGB-D data, there has not been yet any convincing work to eradicate the requirement of in-plane rotated views. Rotation-invariant methods are based on local keypoints in either 2D \cite{Yi2016} or 3D \cite{Drost2010,Birdal2015,Tombari2010} by explicitly computing or voting for an orientation or a local reference frame, but they fail for objects of poor geometry or texture.

Although rarely mentioned, all of the view-based methods cover only a very small, predefined 6D pose space. Placing the object differently, e.g. on its head, would lead to failure if this view had not been specifically included during training. Unfortunately, additional views increase computation and add to overall ambiguity in the matching stage. Even worse, for all discussed methods, scene sampling is crucial. If too coarse, objects of smaller scale can be missed whereas a fine-grained sampling increases computation and often leads to more false positive detections. Therefore, we explore a path similar to works on large-scale classification where dense feature maps on multiple scales have produced state-of-the-art results. Instead of relying on classifying proposed bounding boxes \cite{Girshick, He2015, Lin2016}, whose performance hinges on the proposals' quality, recent single-shot detectors \cite{Redmon2016, Liu2016} classify a (large) discrete set of fixed bounding boxes. This streamlines the network architecture and gives freedom to the a-priori placement of boxes. 

As for works regressing the pose from RGB images, the related works of \cite{Poirson2016,Mousavian2016} recently extended SSD to include pose estimates for categories. \cite{Mousavian2016} infers 3D bounding boxes of objects in urban traffic and regresses 3D box corners and an azimuth angle whereas \cite{Poirson2016} introduces an additional binning of poses to express not only the category but also a notion of local orientation such as 'bike from the side' or 'plane from below'. The difference to us is that they train on real images to predict poses in a very constrained subspace. Instead, our domain demands training on synthetic model-based data and the need to encompass the full 6D pose space to accomplish tasks such as grasping or AR. 

\section{Methodology}
The input to our method is an RGB image that is processed by the network to output localized 2D detections with bounding boxes. Additionally, each 2D box is provided with a pool of the most likely 6D poses for that instance. To represent a 6D pose, we parse the scores for viewpoint and in-plane rotation that have been inferred from the network and use projective properties to instantiate 6D hypotheses. In a final step, we refine each pose in every pool and select the best after verification. This last step can either be conducted in 2D or optionally in 3D if depth data is available. We present each part now in more detail. 

\subsection{Network architecture}
Our base network is derived from a pre-trained InceptionV4 instance \cite{Szegedy2016} and is fed with a color image (resized to $299 \times 299$) to compute feature maps at multiple scales. In order to get our first feature map of dimensionality $71 \times 71 \times 384$, we branch off before the last pooling layer within the stem and append one 'Inception-A' block. Thereafter, we successively branch off after the 'Inception-A' blocks for a $35 \times 35 \times 384$ feature map, after the 'Inception-B' blocks for a $17 \times 17 \times 1024$ feature map and after the 'Inception-C' blocks for a $9 \times 9 \times 1536$ map.\footnote{We changed the padding of Inception-B  s.t. the next block contains a map with odd dimensionality to always contain a central position.}
To cover objects at larger scale, we extend the network with two more parts. First, a 'Reduction-B' followed by two 'Inception-C' blocks to output a $5 \times 5 \times 1024$ map. Second, one 'Reduction-B' and one 'Inception-C' to produce a $3 \times 3 \times 1024$ map.

From here we follow the paradigm of SSD. Specifically, each of these six feature maps is convolved with prediction kernels that are supposed to regress localized detections from feature map positions. Let $(w_s, h_s, c_s)$ be the width, height and channel depth at scale $s$. For each scale, we train a $3 \times 3 \times c_s$ kernel that provides for each feature map location the scores for object ID, discrete viewpoint and in-plane rotation. Since we introduce a discretization error by this grid, we create $B_s$ bounding boxes at each location with different aspect ratios. Additionally, we regress a refinement of their four corners. If $C, V, R$ are the numbers of object classes, sampled viewpoints and in-plane rotations respectively, we produce a $(w_s, h_s, B_s \times (C +V + R + 4) )$ detection map for the scale $s$. The network has a total number of 21222 possible bounding boxes in different shapes and sizes. While this might seem high, the actual runtime of our method is remarkably low thanks to the fully-convolutional design and the good true negative behavior, which tend to yield a very confident and small set of detections. We refer to Figure \ref{fig:network} for a schematic overview.

\paragraph{Viewpoint scoring versus pose regression} The choice of viewpoint classification over pose regression is deliberate. Although works that do direct rotation regression exist \cite{Kendall2015, Tan2015}, early experimentation showed clearly that the classification approach is more reliable for the task of detecting poses. In particular, it seems that the layers do a better job at scoring discrete viewpoints than at outputting numerically accurate translations and rotations. The decomposition of a 6D pose in viewpoint and in-plane rotation is elegant and allows us to tackle the problem more naturally. While a new viewpoint exhibits a new visual structure, an in-plane rotated view is a non-linear transformation of the same view. Furthermore, simultaneous scoring of all views allows us to parse multiple detections at a given image location, \eg by accepting all viewpoints above a certain threshold. Equally important, this approach allows us to deal with symmetries or views of similar appearance in a straight-forward fashion. 

\subsection{Training stage}

\begin{figure}
	\subfloat{\includegraphics[width=4cm]{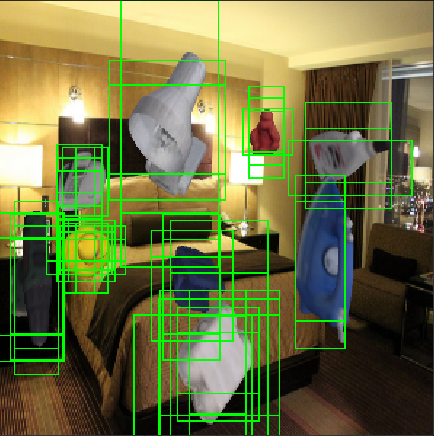}}
	\subfloat{\includegraphics[width=4cm]{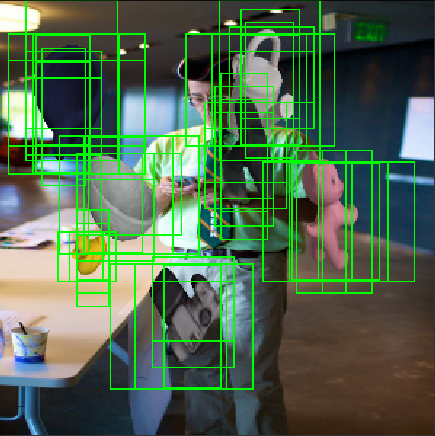}} \\
	\vspace*{-0.4cm}
	\subfloat{\includegraphics[width=4cm]{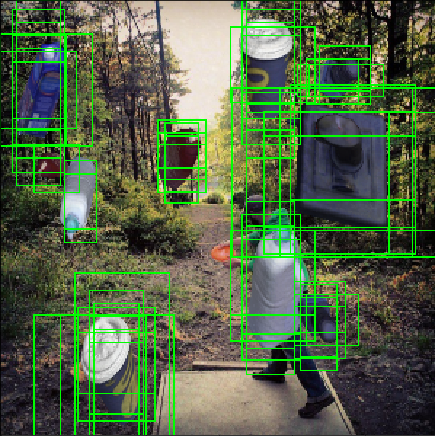}}
	\subfloat{\includegraphics[width=4cm]{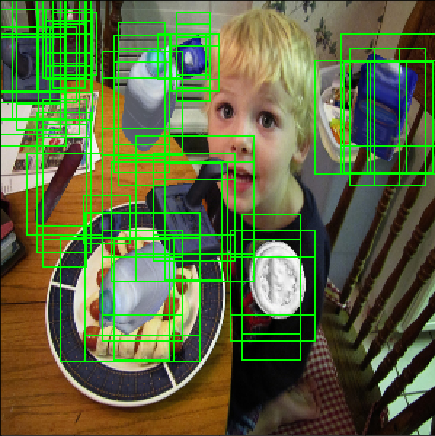}}
	\caption{Exemplary training images for the datasets used. Using MS COCO images as background, we render object instances with random poses into the scene. The green boxes visualize the network's bounding boxes that have been assigned as positive samples for training.}
	\label{fig:training}
\end{figure}

We take random images from MS COCO \cite{Lin2014} as background and render our objects with random transformations into the scene using OpenGL commands. For each rendered instance, we compute the IoU (intersection over union) of each box with the rendered mask and every box $b$ with IoU $> 0.5$ is taken as a positive sample for this object class. Additionally, we determine for the used transformation its closest sampled discrete viewpoint and in-plane rotation as well as set its four corner values to the tightest fit around the mask as a regression target. We show some training images in Figure \ref{fig:training}.

Similar to SSD \cite{Liu2016}, we employ many different kinds of augmentation, such as changing the brightness and contrast of the image. Differently to them, though, we do not flip the images since it would lead to confusion between views and to wrong pose detections later on. We also make sure that each training image contains a 1:2 positives-negatives ratio by selecting hard negatives (unassigned boxes with high object probability) during back-propagation.

Our loss is similar to the MultiBox loss of SSD or YOLO, but we extend the formulation to take discrete views and in-plane rotations into account. Given a set of positive boxes $Pos$ and hard-mined negative boxes $Neg$ for a training image, we minimize the following energy:
\begin{eqnarray}
\label{eq:loss}
L(Pos, Neg) := \sum_{b \in Neg}  L_{class}  + \hspace{1.8cm} \nonumber \\ \sum_{b \in Pos} \left( L_{class} +  \alpha  L_{fit} + \beta L_{view} + \gamma L_{inplane} \right) 
\end{eqnarray}
\begin{figure}
	\includegraphics[width=8cm]{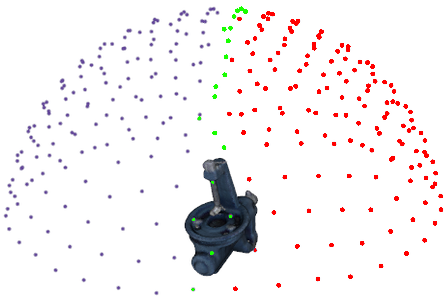}
	\caption{Discrete 6D pose space with each point representing a classifiable viewpoint. If symmetric, we use only the green points for view ID assignment during training whereas semi-symmetric objects use the red points as well.}
	\label{fig:sphere}
\end{figure}

As it can be seen from (\ref{eq:loss}), we sum over positive and negative boxes for class probabilities ($L_{class}$). Additionally, each positive box contributes weighted terms for viewpoint ($L_{view}$) and in-plane classification ($L_{inplane}$), as well as a fitting error of the boxes' corners ($L_{fit}$). For the classification terms, i.e., $L_{class}$, $L_{view}$, $L_{inplane}$, we employ a standard softmax cross-entropy loss, whereas a more robust smooth L1-norm is used for corner regression ($L_{fit}$).

\paragraph{Dealing with symmetry and view ambiguity}
Our approach demands the elimination of viewpoint confusion for proper convergence. We thus have to treat symmetrical or semi-symmetrical (constructible with plane reflection) objects with special care. Given an equidistantly-sampled sphere from which we take our viewpoints, we discard positions that lead to ambiguity. For symmetric objects, we solely sample views along an arc, whereas for semi-symmetric objects we omit one hemisphere entirely. This approach easily generalizes to cope with views which are mutually indistinguishable although this might require manual annotation for specific objects in practice. In essence, we simply ignore certain views from the output of the convolutional classifiers during testing and take special care of viewpoint assignment in training. We refer to Figure \ref{fig:sphere} for a visualization of the pose space.

\subsection{Detection stage}
\begin{figure}
	\subfloat{\includegraphics[width=4cm]{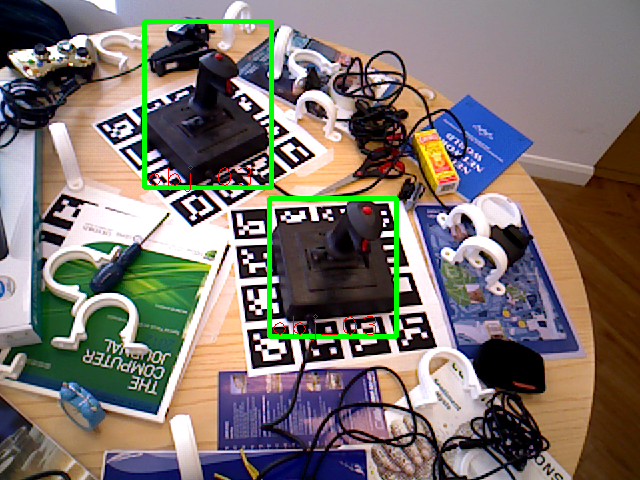}}
	\subfloat{\includegraphics[width=4cm]{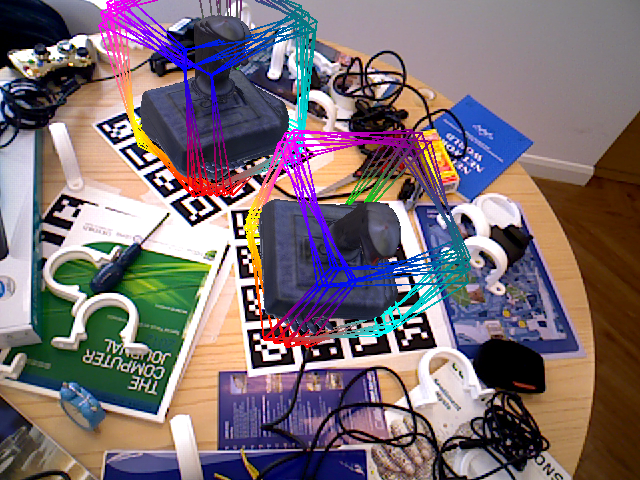}} \\
	\vspace*{-0.4cm}
	\subfloat{\includegraphics[width=4cm]{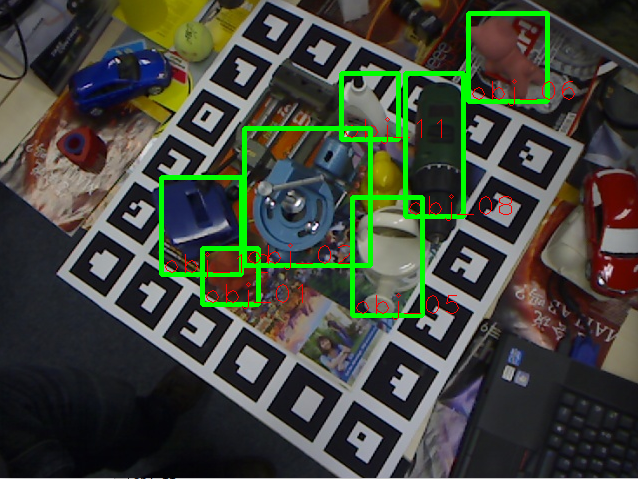}}
	\subfloat{\includegraphics[width=4cm]{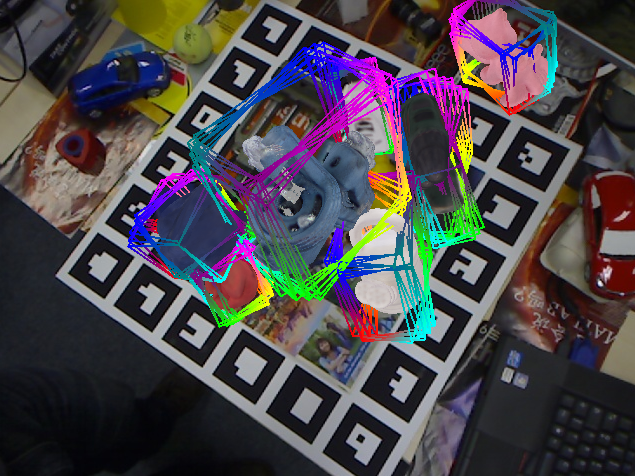}}
	\caption{Prediction output and 6D pose pooling of our network on the Tejani dataset and the multi-object dataset. Each 2D prediction builds a pool of 6D poses by parsing the most confident views and in-plane rotations. Since our networks are trained with various augmentations, they can adapt to different global illumination settings.}
	\label{fig:pool}
\end{figure}

We run a forward-pass on the input image to collect all detections above a certain threshold, followed by non-maximum suppression. This yields refined and tight 2D bounding boxes with an associated object ID and scores for all views and in-plane rotations. For each detected 2D box we thus parse the most confident views as well as in-plane rotations to build a pool of 6D hypotheses from which we select the best after refinement. See Figure \ref{fig:pool} for the pooled hypotheses and Figure \ref{fig:refinement} for the final output.

\subsubsection{From 2D bounding box to 6D hypothesis }
\begin{figure}
	\includegraphics[width=8cm]{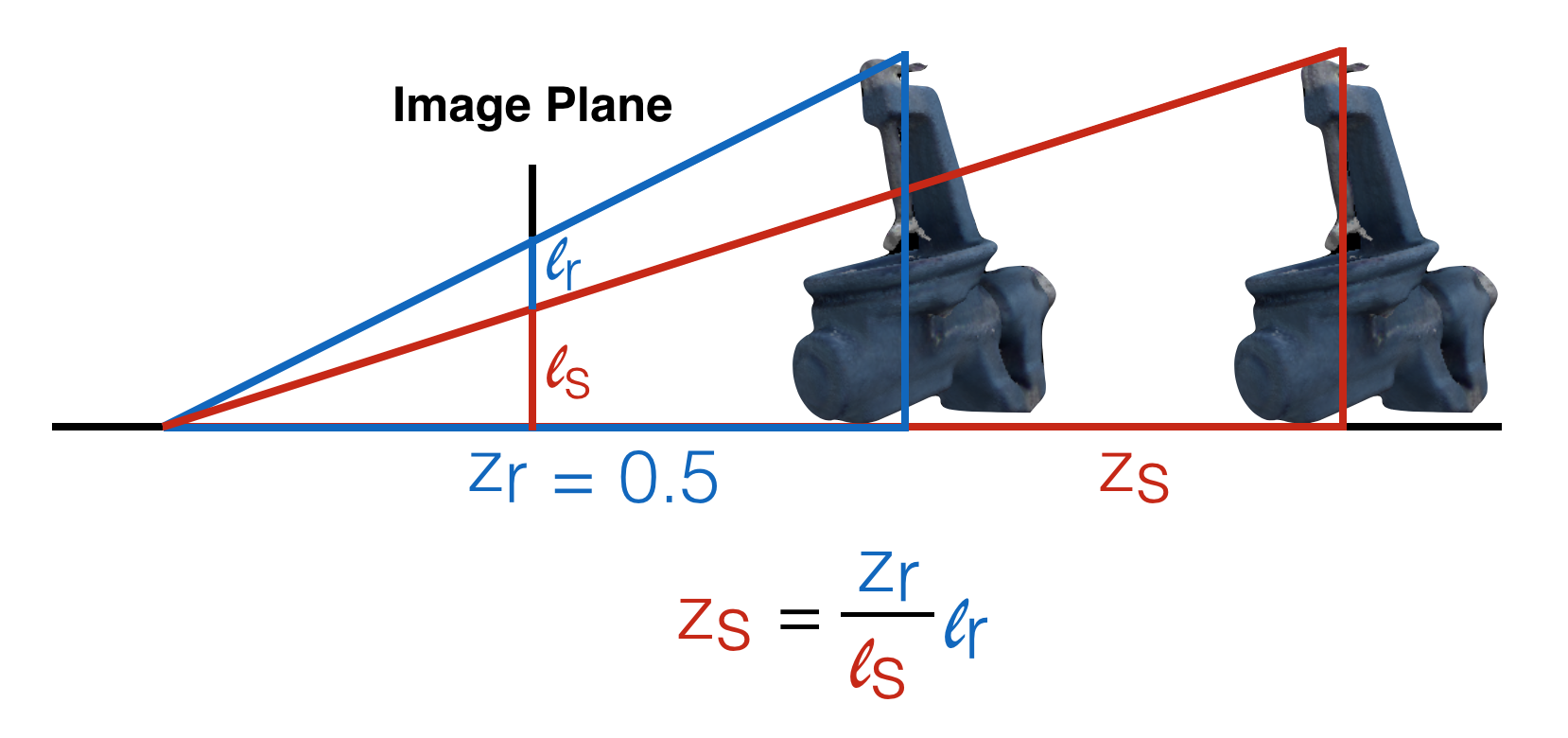}
	\caption{For each object we precomputed the perfect bounding box and the 2D object centroid with respect to each possible discrete rotation in a prior offline stage. To this end, we rendered the object at a canonical centroid distance $z_r=0.5m$. Subsequently, the object distance $z_s$ can be inferred from the projective ratio according to $z_s = \frac{l_r}{l_s} z_r$, where $l_r$ denotes diagonal length of the precomputed bounding box and $l_s$ denotes the diagonal length of the predicted bounding box on the image plane.}
	\label{fig:translation}
\end{figure}

So far, all computation has been conducted on the image plane and we need to find a way to hypothesize 6D poses from our network output. We can easily construct a 3D rotation, given view ID and in-plane rotation ID, and can use the bounding box to infer 3D translation. To this end, we render all possible combinations of discrete views and in-plane rotations at a canonical centroid distance $z_r=0.5m$ in an offline stage and compute their bounding boxes. Given the diagonal length $l_r$ of the bounding box during this offline stage and the one predicted by the network $l_r$, we can infer the object distance $z_s = \frac{l_r}{l_s} z_r$ from their projective ratio, as illustrated in Figure \ref{fig:translation}. In a similar fashion, we can derive the projected centroid position and back-project to a 3D point with known camera intrinsics.

\subsubsection{Pose refinement and verification}
The obtained poses are already quite accurate, yet can in general benefit from a further refinement. Since we will regard the problem for both RGB and RGB-D data, the pose refinement will either be done with an edge-based or cloud-based ICP approach. If using RGB only, we render each hypothesis into the scene and extract a sparse set of 3D contour points. Each 3D point $X_i$, projected to $\pi(X_i)=x_i$, then shoots a ray perpendicular to its orientation to find the closest scene edge $y_i$. We seek the best alignment of the 3D model such that the average projected error is minimal:
\begin{equation}
\argmin _{R,t} \sum_i \bigg( || \pi(R \cdot X_i + t) - y_i || ^2 \bigg).
\end{equation}

We minimize this energy with an IRLS approach (similar to \cite{Drummond2002}) and robustify it using Geman-McLure weighting. In the case of RGB-D, we render the current pose and solve with standard projective ICP with a point-to-plane formulation in closed form \cite{Besl1992}. In both cases, we run multiple rounds of correspondence search to improve refinement and we use multi-threading to accelerate the process.

The above procedure provides multiple refined poses for each 2D box and we need to choose the best one. To this end, we employ a verification procedure. Using only RGB, we do a final rendering and compute the average deviation of orientation between contour gradients and overlapping scene gradients via absolute dot products. In case RGB-D data is available, we render the hypotheses and estimate camera-space normals to measure the similarity again with absolute dot products.

\begin{figure*}	
	\subfloat[2D Detections]{\includegraphics[width=4.3cm]{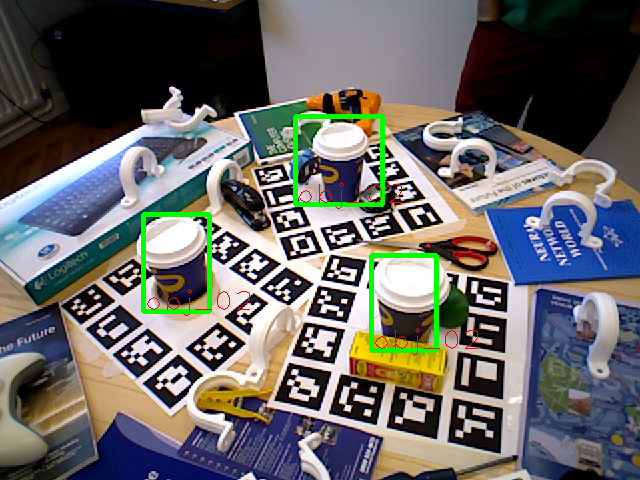}}
	\hspace{0.001cm}
	\subfloat[Unrefined]{\includegraphics[width=4.3cm]{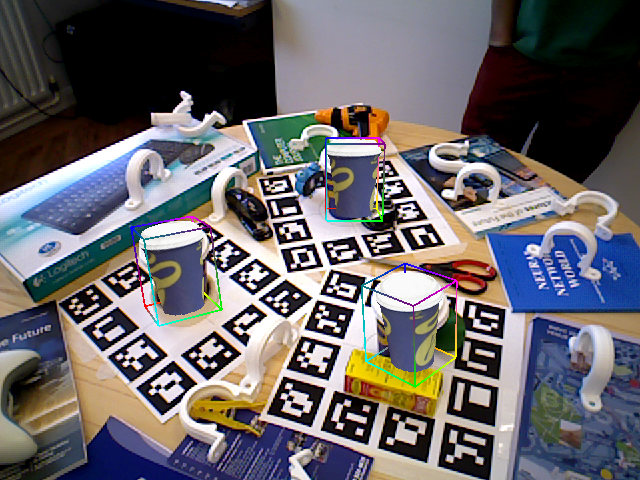}}
	\hspace{0.001cm}
	\subfloat[RGB refinement]{\includegraphics[width=4.3cm]{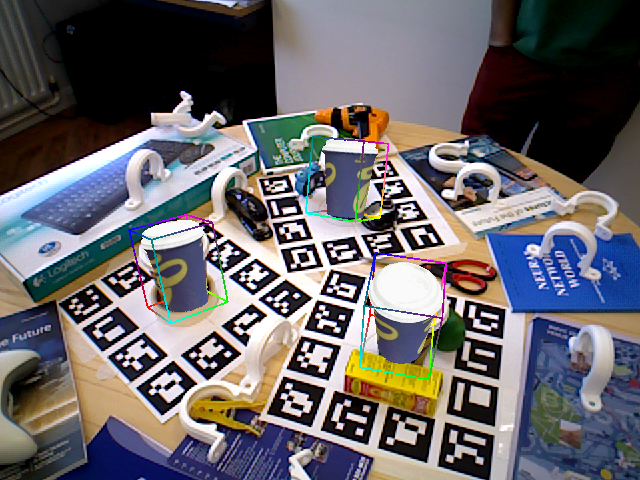}}
	\hspace{0.001cm}
	\subfloat[RGB-D refinement]{\includegraphics[width=4.3cm]{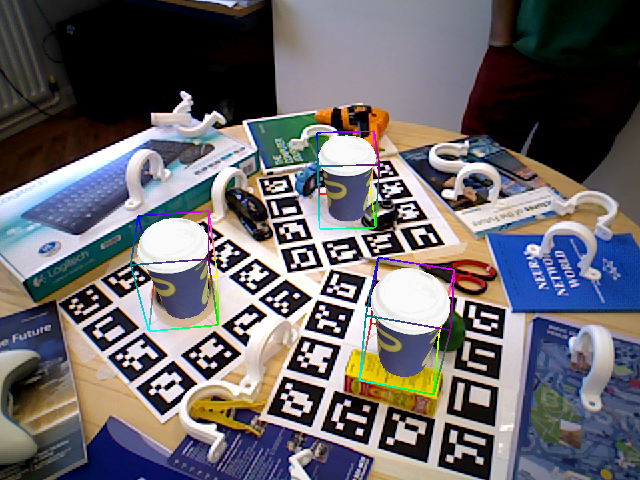}}
	\caption{After predicting 2D detections (a), we build 6D hypotheses and run pose refinement and a final verification. While the unrefined poses (b) are rather approximate, contour-based refinement (c) produces already visually acceptable results. Occlusion-aware projective ICP with cloud data (d) leads to a very accurate alignment.}
	\label{fig:refinement}
\end{figure*}

\section{Evaluation}
We implemented our method in C++ using TensorFlow 1.0 \cite{Abadi2016} and cuDNN 5 and ran it on a i7-5820K@3.3GHz with an NVIDIA GTX 1080. Our evaluation has been conducted on three datasets. The first, presented in Tejani et al. \cite{Tejani2014}, consists of six sequences where each sequence requires the detection and pose estimation of multiple instances of the same object in clutter and with different levels of mild occlusion. The second dataset, presented in \cite{Hinterstoisser2012}, consists of 15 sequences where each frame presents one instance to detect and the main challenge is the high amount of clutter in the scene. As others, we will skip two sequences since they lack a meshed model. The third dataset, presented in \cite{Brachmann2014} is an extension of the second where one sequence has been annotated with instances of multiple objects undergoing heavy occlusions at times.

\paragraph{Network configuration and training}
To get the best results it is necessary to find an appropriate sampling of the model view space. If the sampling is too coarse we either miss an object in certain poses or build suboptimal 6D hypotheses whereas a very fine sampling can lead to a more difficult training. We found an equidistant sampling of the unit sphere into 642 views to work well in practice. Since the datasets only exhibit the upper hemisphere of the objects, we ended up with 337 possible view IDs. Additionally, we sampled the in-plane rotations from -45 to 45 degrees in steps of 5 to have a total of 19 bins.

Given the above configuration, we trained the last layers of the network and the predictor kernels using ADAM and a constant learning rate of $0.0003$ until we saw convergence on a synthetic validation set. The balancing of the loss term weights proved to be vital to provide both good detections and poses. After multiple trials we determined  $\alpha=1.5$, $\beta=2.5$ and $\gamma=1.5$ to work well for us. We refer the reader to the supplementary material to see the error development for different configurations.

\subsection{Single object scenario}

\begin{table}
	\scalebox{0.9}{
		\begin{tabular}{c|c|c|c|c}
			Sequence		& LineMOD \cite{Hinterstoisser2012a} & LC-HF  \cite{Tejani2014}	& Kehl \cite{Kehl2016a} & Us \\
			\hline	    
			Camera 	& 0.589 & 0.394 & 	0.383 & \textbf{0.741} \\
			Coffee   	& 0.942 & 0.891 &0.972 & 	\textbf{0.983} \\
			Joystick   & 0.846 & 0.549	& 0.892 & \textbf{0.997}\\
			Juice 	& 0.595	& 0.883 & 0.866 & \textbf{0.919} \\
			Milk 		& 0.558 & 0.397	& 0.463 & \textbf{0.780} \\
			Shampoo	& \textbf{0.922}	& 0.792	& 0.910 & 0.892 \\
			\hline
			Total	&  0.740     & 0.651	& 0.747 & \textbf{0.885}\\
		\end{tabular}
		\caption{F1-scores on the re-annotated version of \cite{Tejani2014}. Although our method is the only one to solely use RGB data, our results are considerably higher than all related works.}
		\label{table:tejani_f1}	}
\end{table}	

\begin{table*}
	\begin{center}
		\begin{tabular}{@{}c|c|c|c|c|c|c|c|c|c|c|c|c|c@{}}
			& ape & bvise   & cam & can & cat  & driller & duck & box & glue & holep & iron & lamp & phone \\ \hline
			Us  & 76.3 & \textbf{97.1}  &  92.2 & \textbf{93.1} & 89.3 & \textbf{97.8} & 80.0 & 93.6 & \textbf{76.3} & 71.6	  & \textbf{98.2} & 93.0 & \textbf{92.4}     \\
			LineMOD \cite{Hinterstoisser2012} & 53.3 & 84.6   & 64.0 & 51.2 & 65.6 &  69.1 & 58.0 & 86.0 & 43.8 & 51.6 & 68.3 & 67.5 & 56.3 \\
			LC-HF \cite{Tejani2014} & 85.5 		& 96.1  &  71.8 & 70.9 & 88.8 &  90.5 & 90.7 & 74.0 & 67.8 & 87.5 & 73.5 & 92.1 & 72.8 \\
			Kehl \cite{Kehl2016a}    & \textbf{98.1}& 94.8    &  \textbf{93.4} & 82.6 & \textbf{98.1} &  96.5 & \textbf{97.9} & \textbf{100}  & 74.1 & \textbf{97.9}  & 91.0 & \textbf{98.2} & 84.9     \\
			
		\end{tabular}
	\end{center}
	\caption{F1-scores for each sequence of \cite{Hinterstoisser2012}. Note that the LineMOD scores are supplied from \cite{Tejani2014} with their evaluation since \cite{Hinterstoisser2012} does not provide them. Using color only we can easily compete with the other RGB-D based methods.}
	\label{table:linemod_f1}
\end{table*}

Since 3D detection is a multi-stage pipeline for us, we first evaluate purely the 2D detection performance between our predicted boxes and the tight bounding boxes of the rendered groundtruth instances on the first two datasets. Note that we always conduct proper detection and not localization, \ie we do not constrain the maximum number of allowed detections but instead accept all predictions above a chosen threshold. We count a detection to be correct when the IoU score of a predicted bounding box with the groundtruth box is higher than 0.5. We present our F1-scores in Tables \ref{table:tejani_f1} and \ref{table:linemod_f1} for different detection thresholds. 

It is important to mention that the compared methods, which all use RGB-D data, allow a detection to survive after rigorous color- and depth-based checks whereas we use simple thresholding for each prediction. Therefore, it is easier for them to suppress false positives to increase their precision whereas our confidence comes from color cues only.

On the Tejani dataset we outperform all related RGB-D methods by a huge margin of $13.8 \%$ while using color only. We analyzed the detection quality on the two most difficult sequences. The 'camera' has instances of smaller scale which are partially occluded and therefore simply missed whereas the 'milk' sequence exhibits stronger occlusions in virtually every frame. Although we were able to detect the 'milk' instances, our predictors could not overcome the occlusions and regressed wrongly-sized boxes which were not tight enough to satisfy the IoU threshold. These were counted as false positives and thus lowered our recall\footnote{We refer to the supplement for more detailed graphs.}.

On the second dataset we have mixed results where we can outperform state-of-the-art RGB-D methods on some sequences while being worse on others. For larger feature-rich objects like 'benchvise', 'iron' or 'driller' our method performs better than the related work since our network can draw from color and textural information.
For some objects, such as 'lamp' or 'cam', the performance is worse than the related work. Our method relies on color information only and thus requires a certain color similarity between synthetic renderings of the CAD model and their appearance in the scene. Some objects exhibit specular effects (\ie changing colors for different camera positions) or the frames can undergo sensor-side changes of exposure or white balancing, causing a color shift. Brachmann et al. \cite{Brachmann2016} avoid this problem by training on a well-distributed subset of real sequence images. Our problem is much harder since we train on synthetic data only and must generalize to real, unseen imagery.

Our performance for objects of smaller scale such as  'ape', 'duck' and 'cat' is worse and we observed a drop both in recall and precision. We attribute the lower recall to our bounding box placement, which can have 'blind spots' at some locations and consequently, leading to situations where a small-scale instance cannot be covered sufficiently by any box to fire. The lower precision, on the other hand, stems from the fact that these objects are textureless and of uniform color which increases confusion with the heavy scene clutter.

\subsubsection{Pose estimation}
We chose for each object the threshold that yielded the highest F1-score and run all following pose estimation experiments with this setting. We are interested in the pose accuracy for all correctly detected instances.

\begin{table}
	\scalebox{1.0}{
		\begin{tabular}{c|c|c|c|c}
			Sequence		& IoU-2D & IoU-3D & VSS-2D & VSS-3D \\
			\hline	    
			Camera 	& 0.973 & 0.904 & 	0.693 & 0.778 \\
			Coffee   	& 0.998 & 0.996 &0.765 & 0.931 \\
			Joystick   & 1 & 0.953	& 0.655 & 0.866\\
			Juice 	& 0.994	& 0.962 & 0.742 & 0.865 \\
			Milk 		& 0.970 & 0.990	& 0.722 & 0.810 \\
			Shampoo	& 0.993	& 0.974	& 0.767 & 0.874 \\
			\hline
			Total	&  0.988     & 0.963	& 0.724 & 0.854 \\
		\end{tabular}
		\caption{Average pose errors for the Tejani dataset.}
		\label{table:tejani_pose}		
	}
\end{table}

\begin{table}
	\scalebox{0.85}{
		\begin{tabular}{c|c|c|c}
			& \multicolumn{3}{c}{RGB} \\
			& Us & LineMOD\cite{Hinterstoisser2011} &  Brachmann \cite{Brachmann2016}  \\
			\hline
			IoU & 99.4 \% & 86.5\% & 97.5\% \\
			ADD & 76.3\% & 24.2\% & 50.2\%  \\
		\end{tabular}
	}
\end{table}
\begin{table}
	\scalebox{0.85}{
		\vspace*{5mm}
		\begin{tabular}{c|c|c|c}
			& \multicolumn{3}{c}{RGB-D} \\
			& Ours & Brachmann 2016 \cite{Brachmann2016} & Brachmann 2014 \cite{Brachmann2014} \\
			\hline
			IoU & 96.5 \% & 99.6\% & 99.1\% \\
			ADD \cite{Hinterstoisser2012a} & 90.9\% & 99.0\% & 97.4\% \\
		\end{tabular}
		\caption{Average pose errors for the LineMOD dataset.}
		\label{table:linemod_pose}		
	}
\end{table}	

\paragraph{Error metrics} To measure 2D pose errors we will compute both an IoU score and a Visual Surface Similarity (VSS) \cite{Hodan2016}. The former is different than the detection IoU check since it measures the overlap of the rendered masks' bounding boxes between groundtruth and final pose estimate and accepts a pose if the overlap is larger than $0.5$. VSS is a tighter measure since it counts the average pixel-wise overlap of the mask. This measure assesses well the suitability for AR applications and has the advantage of being agnostic towards the symmetry of objects. To measure the 3D pose error we use the ADD score from \cite{Hinterstoisser2012}. This assesses the accuracy for manipulation tasks by measuring the average deviation between transformed model point clouds of groundtruth and hypothesis. If it is smaller than $\frac{1}{10}th$ of the model diameter, it is counted as a correct pose.

\paragraph{Refinement with different parsing values} 
\begin{figure}
	\includegraphics[width=8cm]{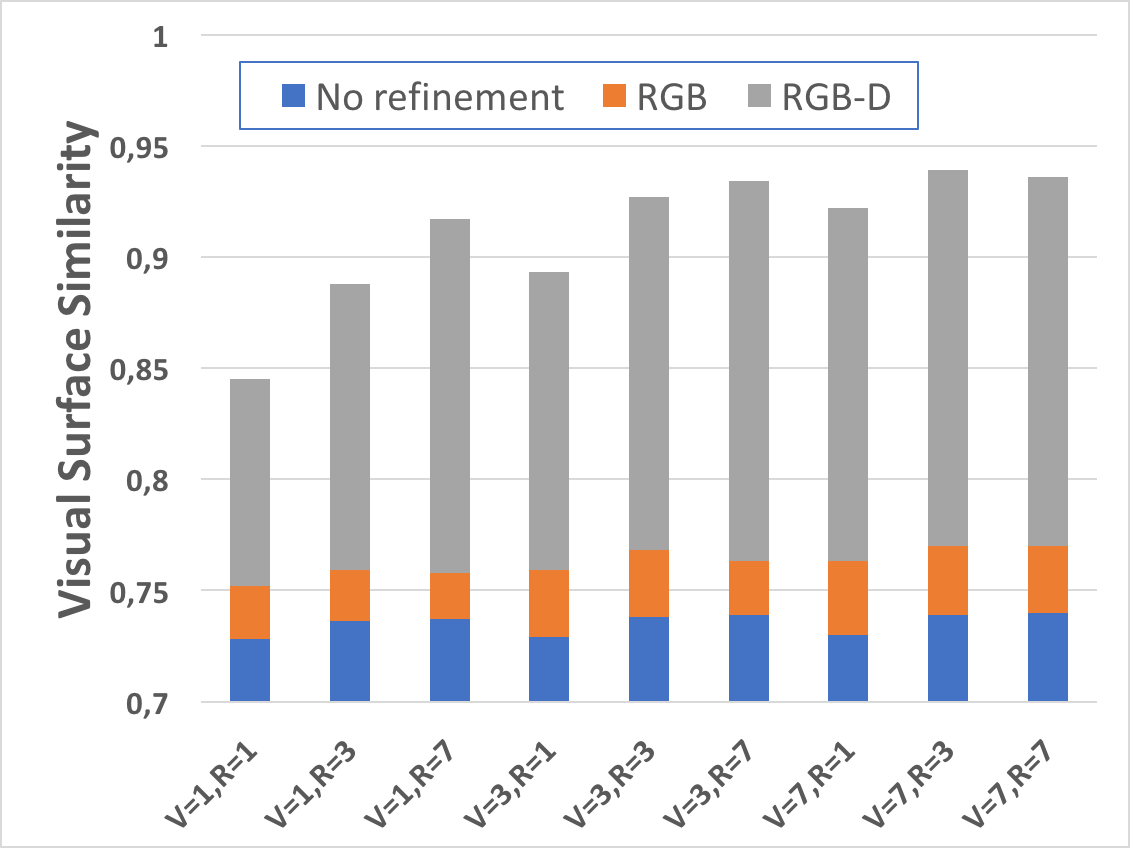}	
	\caption{Average VSS scores for the 'coffee' object for different numbers of parsed views and in-plane rotations as well as different pose refinement options.}
	\label{fig:parsing}
\end{figure}
As mentioned, we parse the most confident views and in-plane rotations to build a pool of 6D hypotheses for each 2D detection. Here, we want to assess the final pose accuracy when changing the number of parsed views $V$ and rotations $R$ for different refinement strategies We present in Figure \ref{fig:parsing} the results on Tejani's 'coffee' sequence for the cases of no refinement, edge-based and cloud-based refinement (see Figure \ref{fig:refinement} for an example). To decide for the best pose we employ verification over contours  for the first two cases and normals for the latter. As can be seen, the final poses without any refinement are imperfect but usually provide very good initializations for further processing. Additional 2D refinement yields better poses but cannot cope well with occluders whereas depth-based refinement leads to perfect poses in practice. The figure gives also insight for varying $V$ and $R$ for hypothesis pool creation. Naturally, with higher numbers the chances of finding a more accurate pose improve since we evaluate a larger portion of the 6D space. It is evident, however, that every additional parsed view $V$ gives a larger benefit than taking more in-plane rotations $R$ into the pool. We explain this by the fact that our viewpoint sampling is coarser than our in-plane sampling and thus reveals more uncovered pose space when parsed, which in turn helps especially depth-based refinement. Since we create a pool of $V \cdot R$ poses for each 2D detection, we fixed $V=3,R=3$ for all experiments as a compromise between accuracy and refinement runtime.

\paragraph{Performance on the two datasets}

We present our pose errors in Tables \ref{table:tejani_pose} and \ref{table:linemod_pose} after 2D and 3D refinement. Note that we do not compute the ADD scores for Tejani since each object is of (semi-)symmetric nature, leading always to near-perfect ADD scores of 1. The poses are visually accurate after 2D refinement and furthermore are boosted by an additional depth-based refinement stage. On the second dataset we are actually able to come very close to Brachmann et al. which is surprising since they have a huge advantage of real data training. For the case of pure RGB-based poses, we can even overtake their results. We provide more detailed error tables in the supplement.

\begin{figure}
	\includegraphics[width=4cm, height=4cm]{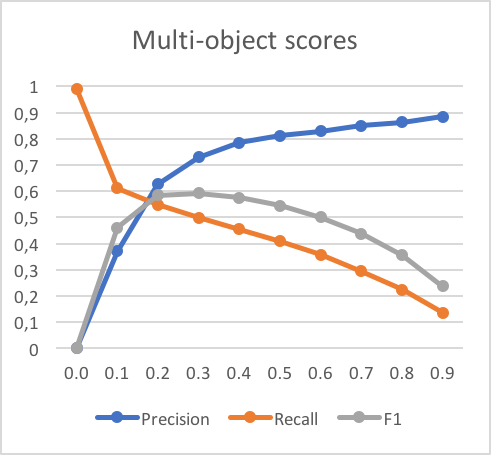}	
	\includegraphics[width=4cm, height=4cm]{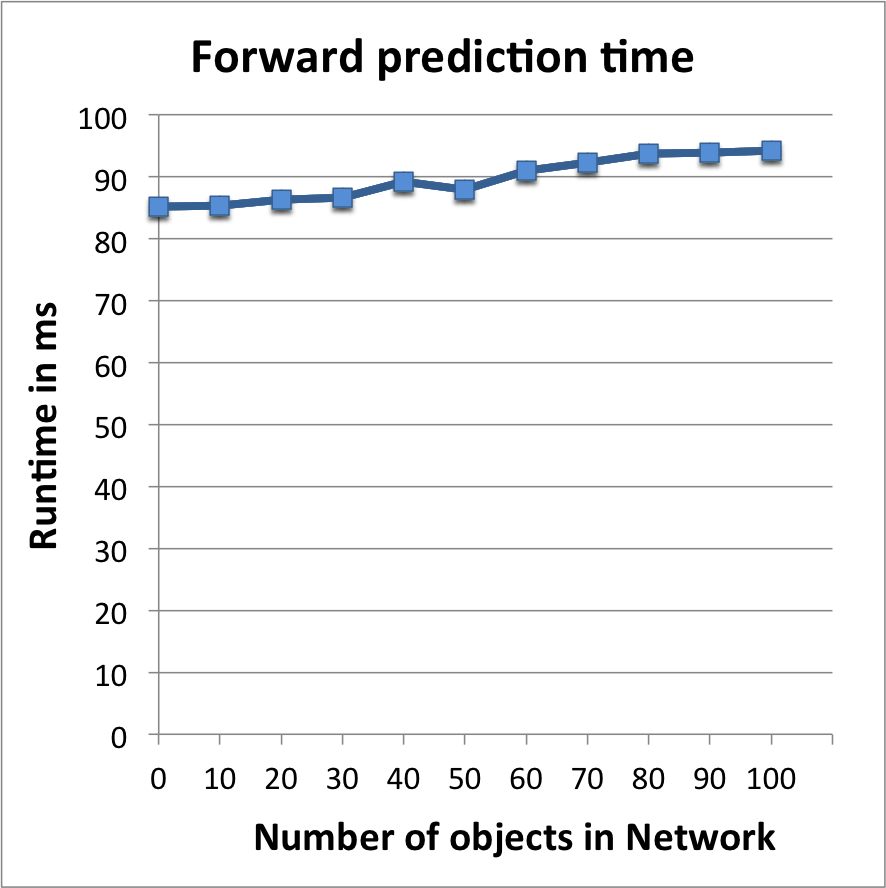}	
	\caption{Left: Detection scores on the multi-object dataset for a different global threshold. Right: Runtime increase for the network prediction with an increased number of objects.}
	\label{fig:multi}
\end{figure}

\subsection{Multiple object detection}
The last dataset has annotations for 9 out of the 15 objects and is quite difficult since many instances undergo heavy occlusion. Different to the single object scenario, we have now a network with one global detection threshold for all objects and we present our scores in Figure \ref{fig:multi} when varying this threshold. Brachmann et al. \cite{Brachmann2016} can report an impressive Average Precision (AP) of 0.51 whereas we can report an AP of 0.38. It can be observed that our method degrades gracefully as the recall does not drop suddenly from one threshold step to the next. Note again that Brachmann et al. have the advantage of training on real images of the sequence whereas we must detect heavily-occluded objects from synthetic training only.

\subsection{Runtime and scalability}
For a single object in the database, Kehl et al. \cite{Kehl2016a} report a runtime of around 650ms per frame whereas Brachmann et al. \cite{Brachmann2014, Brachmann2016}  report around 450ms. Above methods are scalable and thus have a sublinear runtime growth with an increasing database size. Our method is a lot faster than the related work while being scalable as well. In particular, we can report a runtime of approximately 85ms for a single object. We show our prediction times in Figure \ref{fig:multi} which reveals that we scale very well with an increasing number of objects in the network. While the prediction is fast, our pose refinement takes more time since we need to refine every pose of each pool. On average, given that we have about 3 to 5 positive detections per frame, we need a total of an additional 24ms for refinement, leading to a total runtime of around 10Hz.

\begin{figure}
	\includegraphics[width=4cm]{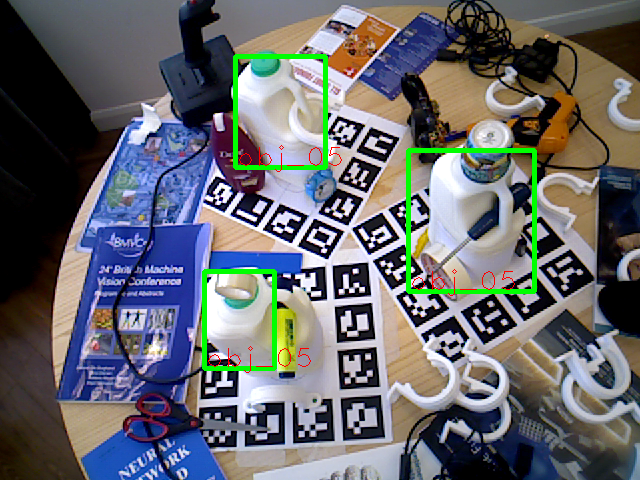}	
	\includegraphics[width=4cm]{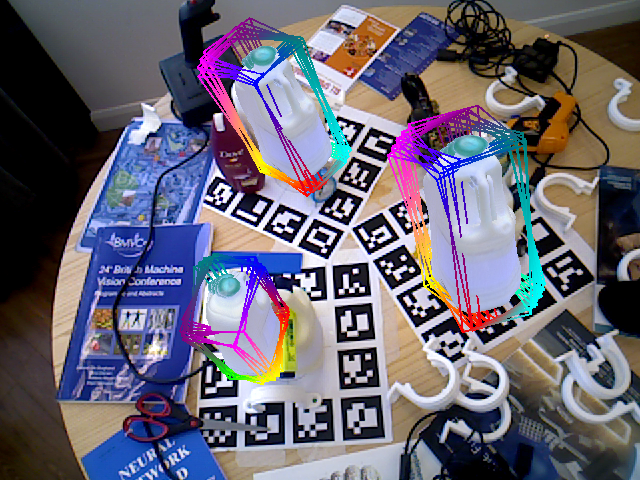}	
	
	\caption{One failure case where incorrect bounding box regression, induced by occlusion, led to wrong 6D hypothesis creation. In such cases a subsequent refinement cannot always recover the correct pose anymore.}
	\label{fig:failure}
\end{figure}

\subsection{Failure cases}
The most prominent issue is the difference in colors between synthetic model and scene appearance, also including local illumination changes such as specular reflections. In these cases, the object confidence might fall under the detection threshold since the difference between the synthetic and the real domain is too large. A more advanced augmentation would be needed to successfully tackle this problem. Another possible problem can stem from the bounding box regression. If the regressed corners are not providing a tight fit, it can lead to translations that are too offset during 6D pose construction. An example of this problem can be seen in Figure \ref{fig:failure} where the occluded milk produces wrong offsets.  We also observed that small objects are sometimes difficult to detect which is even more true after resizing the input to $299 \times 299$. Again, designing a more robust training as well as a larger network input could be of benefit here.

\section*{Conclusion}
To our knowledge, we are the first to present an SSD-style detector for 3D instance detection and full 6D pose estimation that is trained on synthetic model information. We have shown that color-based detectors are indeed able to match and surpass current state-of-the-art methods that leverage RGB-D data while being around one order of magnitude faster.
Future work should include a higher robustness towards color deviation between CAD model and scene appearance. Avoiding the problem of proper loss term balancing is also an interesting direction for future research.

{\small
\bibliographystyle{ieee}
\bibliography{egbib}

\begin{thebibliography}{10}\itemsep=-1pt

\bibitem{Abadi2016}
M.~Abadi, P.~Barham, J.~Chen, Z.~Chen, A.~Davis, J.~Dean, M.~Devin,
  S.~Ghemawat, G.~Irving, M.~Isard, M.~Kudlur, J.~Levenberg, R.~Monga,
  S.~Moore, D.~Murray, B.~Steiner, P.~Tucker, V.~Vasudevan, P.~Warden,
  M.~Wicke, Y.~Yu, and X.~Zheng.
\newblock {TensorFlow: Large-scale machine learning on heterogeneous systems}.
\newblock In {\em OSDI}, 2016.

\bibitem{Besl1992}
P.~Besl and N.~McKay.
\newblock {A Method for Registration of 3-D Shapes}.
\newblock {\em TPAMI}, 1992.

\bibitem{Birdal2015}
T.~Birdal and S.~Ilic.
\newblock {Point Pair Features Based Object Detection and Pose Estimation
  Revisited}.
\newblock In {\em 3DV}, 2015.

\bibitem{Brachmann2014}
E.~Brachmann, A.~Krull, F.~Michel, S.~Gumhold, J.~Shotton, and C.~Rother.
\newblock {Learning 6D Object Pose Estimation using 3D Object Coordinates}.
\newblock In {\em ECCV}, 2014.

\bibitem{Brachmann2016}
E.~Brachmann, F.~Michel, A.~Krull, M.~Y. Yang, S.~Gumhold, and C.~Rother.
\newblock {Uncertainty-Driven 6D Pose Estimation of Objects and Scenes from a
  Single RGB Image}.
\newblock In {\em CVPR}, 2016.

\bibitem{Doumanoglou2016}
A.~Doumanoglou, R.~Kouskouridas, S.~Malassiotis, and T.-K. Kim.
\newblock {6D Object Detection and Next-Best-View Prediction in the Crowd}.
\newblock In {\em CVPR}, 2016.

\bibitem{Drost2010}
B.~Drost, M.~Ulrich, N.~Navab, and S.~Ilic.
\newblock {Model globally, match locally: efﬁcient and robust 3D object
  recognition}.
\newblock In {\em CVPR}, 2010.

\bibitem{Drummond2002}
T.~Drummond and R.~Cipolla.
\newblock {Real-time visual tracking of complex structures}.
\newblock {\em TPAMI}, 2002.

\bibitem{Everingham2014}
M.~Everingham, S.~M.~A. Eslami, L.~{Van Gool}, C.~K.~I. Williams, J.~Winn, and
  A.~Zisserman.
\newblock {The Pascal Visual Object Classes Challenge: A Retrospective}.
\newblock {\em IJCV}, 2014.

\bibitem{Girshick}
R.~Girshick.
\newblock {Fast R-CNN}.
\newblock {\em arXiv:1504.08083}, 2015.

\bibitem{He2015}
K.~He, X.~Zhang, S.~Ren, and J.~Sun.
\newblock {Spatial Pyramid Pooling in Deep Convolutional Networks for Visual
  Recognition}.
\newblock {\em TPAMI}, 2015.

\bibitem{Hinterstoisser2012a}
S.~Hinterstoisser, C.~Cagniart, S.~Ilic, P.~Sturm, N.~Navab, P.~Fua, and
  V.~Lepetit.
\newblock {Gradient Response Maps for Real-Time Detection of Textureless
  Objects}.
\newblock {\em TPAMI}, 2012.

\bibitem{Hinterstoisser2011}
S.~Hinterstoisser, S.~Holzer, C.~Cagniart, S.~Ilic, K.~Konolige, N.~Navab, and
  V.~Lepetit.
\newblock {Multimodal templates for real-time detection of texture-less objects
  in heavily cluttered scenes}.
\newblock In {\em ICCV}, 2011.

\bibitem{Hinterstoisser2012}
S.~Hinterstoisser, V.~Lepetit, S.~Ilic, S.~Holzer, G.~Bradsky, K.~Konolige, and
  N.~Navab.
\newblock {Model Based Training, Detection and Pose Estimation of Texture-Less
  3D Objects in Heavily Cluttered Scenes}.
\newblock In {\em ACCV}, 2012.

\bibitem{Hodan2016}
T.~Hodan, J.~Matas, and S.~Obdrzalek.
\newblock {On Evaluation of 6D Object Pose Estimation}.
\newblock In {\em ECCV Workshop}, 2016.

\bibitem{Hodan2015}
T.~Hodan, X.~Zabulis, M.~Lourakis, S.~Obdrzalek, and J.~Matas.
\newblock {Detection and Fine 3D Pose Estimation of Textureless Objects in
  RGB-D Images}.
\newblock In {\em IROS}, 2015.

\bibitem{Kehl2016a}
W.~Kehl, F.~Milletari, F.~Tombari, S.~Ilic, and N.~Navab.
\newblock {Deep Learning of Local RGB-D Patches for 3D Object Detection and 6D
  Pose Estimation}.
\newblock In {\em ECCV}, 2016.

\bibitem{Kehl2015}
W.~Kehl, F.~Tombari, N.~Navab, S.~Ilic, and V.~Lepetit.
\newblock {Hashmod: A Hashing Method for Scalable 3D Object Detection}.
\newblock In {\em BMVC}, 2015.

\bibitem{Kendall2015}
A.~Kendall, M.~Grimes, and R.~Cipolla.
\newblock {PoseNet: A Convolutional Network for Real-Time 6-DOF Camera
  Relocalization}.
\newblock In {\em ICCV}, 2015.

\bibitem{Lin2014}
G.~Lin, C.~Shen, Q.~Shi, A.~V.~D. Hengel, and D.~Suter.
\newblock {Fast Supervised Hashing with Decision Trees for High-Dimensional
  Data}.
\newblock In {\em CVPR}, 2014.

\bibitem{Lin2016}
T.-Y. Lin, P.~Doll{\'{a}}r, R.~Girshick, K.~He, B.~Hariharan, and S.~Belongie.
\newblock {Feature Pyramid Networks for Object Detection}.
\newblock In {\em arXiv:1612.03144}, 2016.

\bibitem{Liu2016}
W.~Liu, D.~Anguelov, D.~Erhan, C.~Szegedy, S.~Reed, C.-y. Fu, and A.~C. Berg.
\newblock {SSD : Single Shot MultiBox Detector}.
\newblock In {\em ECCV}, 2016.

\bibitem{Mousavian2016}
A.~Mousavian, D.~Anguelov, J.~Flynn, and J.~Kosecka.
\newblock {3D Bounding Box Estimation Using Deep Learning and Geometry}.
\newblock {\em arXiv:1612.00496}, 2016.

\bibitem{Poirson2016}
P.~Poirson, P.~Ammirato, C.-Y. Fu, W.~Liu, J.~Kosecka, and A.~C. Berg.
\newblock {Fast Single Shot Detection and Pose Estimation}.
\newblock In {\em 3DV}, 2016.

\bibitem{Redmon2016}
J.~Redmon, S.~Divvala, R.~Girshick, and A.~Farhadi.
\newblock {You only look once: Unified, real-time object detection}.
\newblock In {\em CVPR}, 2016.

\bibitem{Russakovsky2015}
O.~Russakovsky, J.~Deng, H.~Su, J.~Krause, S.~Satheesh, S.~Ma, Z.~Huang,
  A.~Karpathy, A.~Khosla, M.~Bernstein, A.~C. Berg, and L.~Fei-Fei.
\newblock {ImageNet Large Scale Visual Recognition Challenge}.
\newblock {\em IJCV}, 2015.

\bibitem{Szegedy2016}
C.~Szegedy, S.~Ioffe, and V.~Vanhoucke.
\newblock {Inception-v4, Inception-ResNet and the Impact of Residual
  Connections on Learning}.
\newblock {\em arxiv:1602.07261}, 2016.

\bibitem{Tan2015}
D.~J. Tan, F.~Tombari, S.~Ilic, and N.~Navab.
\newblock {A Versatile Learning-based 3D Temporal Tracker : Scalable , Robust ,
  Online}.
\newblock In {\em ICCV}, 2015.

\bibitem{Tejani2014}
A.~Tejani, D.~Tang, R.~Kouskouridas, and T.-k. Kim.
\newblock {Latent-class hough forests for 3D object detection and pose
  estimation}.
\newblock In {\em ECCV}, 2014.

\bibitem{Tombari2010}
F.~Tombari, S.~Salti, and L.~{Di Stefano}.
\newblock {Unique signatures of histograms for local surface description}.
\newblock In {\em ECCV}, 2010.

\bibitem{Ulrich2012}
M.~Ulrich, C.~Wiedemann, and C.~Steger.
\newblock {Combining scale-space and similarity-based aspect graphs for fast 3D
  object recognition}.
\newblock {\em TPAMI}, 2012.

\bibitem{Yi2016}
K.~M. Yi, E.~Trulls, V.~Lepetit, and P.~Fua.
\newblock {LIFT: Learned invariant feature transform}.
\newblock In {\em ECCV}, 2016.

\end{thebibliography}
}

\end{document}